# Objective Probability*

## Henry E. Kyburg, Jr.

A distinction is sometimes made between "statistical" and "subjective" probabilities. This is based on a distinction between "unique" events and "repeatable" events. We argue that this distinction is untenable, since all events are "unique" and all events belong to "kinds", and offer a conception of probability for AI in which (1) all probabilities are based on -- possibly vague -- statistical knowledge, and (2) every statement in the language has a probability. This conception of probability can be applied to very rich languages.

1.  <u>Objectivity</u>

When is a probability 'objective'? To say that a certain kind of event has a certain probability in a reference class is to make a claim that is objective, in the sense that it is completely independent of what anyone knows or believes, completely independent of what the evidence supports. The kind of event has that probability or relative frequency in that class, or it doesn't. That is a matter ofjectively determined by the world.

However, the very fact that relative frequencies or limiting frequencies have the values they have, regardless of what anyone knows or believes, suggests that this may not be the most useful concept of probability for making decisions or representing the impact of evidence. Furthermore, as has been pointed out by those who endorse this conception of probability, this notion does *not* apply to the single case: One can speak of the probability of heads in the class of coin tosses, but not the probability of heads on the next toss.

One response to this has been to move to the opposite extreme: the notion of subjective probability. According to this idea, probability represents the *degree of belief* of an individual. Of course for an individual's degrees of belief to be representable by a probability, they must conform to the axioms of the probability calculus. The condition that they so conform, a condition we would naturally impose on any machine representation of beliefs, is construed as a condition of rationality. But given that this condition of rationality is satisfied, given that the degrees in related events are related as the axioms of the probability calculus stipulate, *there are no further constraints to be* *satisfied.* Any distribution of belief is as good as any other.

But there are other sorts of objectivity one can consider. Carnap's (1951) interpretation of probability was perfectly objective, but intended to serve the functions of the subjective interpretation. The idea was that each sentence of a formal language was to receive a measure, and that the probability of a sentence $h$, relative to a body of evidence $e$, $c(h,e)$, was to be defined in terms of that measure. The probability of a sentence $h$ was the value of the probability of the sentence $h$ relative to the total body of evidence of the agent.

This is *not* subjective probability, though probabilities are in a sense relativized to agents. The reason that it is not subjective is that it is the *evidence* and not the individual, that determines the values of the probabilities. *Any* individual having that evidence ought, if he is rational, to have exactly that probability. Probability is not determined by the whim of the individual, but by the *evidence* the individual has at his disposal.

The treatment of probability offered here is objective in this sense. The value of a probability depends on the evidence the individual has at his disposal but not on the accidental interests of the agent. Probability is objective in the sense that logic is: given a set of premises, what is derivable is determined by the laws of logic, not by the interests of an individual who may accept those premises. What implies what is an objective matter. What makes what probable is equally an objective matter.

Finally, the treatment offered here is intended to be monolithic; we intend that this computable concept of probability is the only measure of uncertainty needed in updating knowledge and making decisions.

2. <u>Repeatability</u>

Almost everyone will agree that when our background statistical knowledge is extensive enough, <u>and</u> when the case with which we are concerned is a "repeatable event", then objective probabilities are appropriate, and these are the probabilities that should enter into the computation of expectation and into our decision theory. Unique events are alleged to call for subjective probability. But I claim that from a down-to-earth <u>practical</u> point of view, from the point of view that seeks to compute probabilities and
expectations for making decisions, the distinction between
'repeatable' and 'unique' events is not only untenable, but
seriously misleading.

The classical repeatable event is the toss of a coin.
Not only can we toss a coin over and over again; coins *have*
been tossed over and over again, and in the experience of
each of us there is a large database of results of coin tosses
(or an impressionistic resume of such a database). And so we
can regard <u>the next toss</u> as a member of a class of tosses, of
which we have reason to believe that half land heads.

But not <u>the last toss</u>. The fourteenth toss? It
depends on what we know about it. Has it been performed?
Do we actually know its result? A particular toss, at a
particular time and place, cannot be repeated. We all know
that. But the event *can* be repeated in 'all relevant
respects'. We don't have to make the toss at the same time
or the same place; we don't have to use the same coin; we
don't have to use the same kind of coin; we don't have to
flip it in any particular way.

These are matters that we judge to be the case.
Does that mean that 'repeatability' is a subjective matter?
Not at all. We learn what the relevant factors are; it is a
matter of objective knowledge.

In principle, many people would say, we could
perfectly well know enough to predict the outcome of a single
specified toss. If we knew the momenta imparted to the
coin, its distance from the surface on which it is to land,
the relevant coefficients of elasticity, etc., we could
presumably predict the outcome of the toss. Every toss is
unique with respect to these properties. But that does not
prevent us from regarding ordinary coin tosses (correctly) as
repeatable events.

3. <u>Uniqueness</u>

Louis Narens [1985, 282] argues that "the
uncertainties that people encounter in the everyday world,
that businessmen encounter in their economic activities, that
the military encounters in war, etc., are not of this
[repeatable] type: repeatability of "experiments" are [sic] not
feasible and often impossible."

Suppose I am interested in whether or not my friend

Sam will be at home tonight after supper. Whether or not he will be home depends on factors I can specify. If there is a good Western in town he may be more likely to be at the movies. If there were to be a chemistry examination tomorrow, Sam would most likely be at home. In short, this seems like a perfect case for subjective probability. But if I have known Sam for a long time, I *do* have a basis for knowing how often, in general, he goes to the movies on week-nights. My knowledge is neither so precise nor so secure as my knowledge about the coin, but it is surely not non-existent.

We must beware of allowing the variety of our knowledge about Sam to serve as an excuse for guessing wildly. Analogously, if we were to have detailed and microscopic data concerning the coin toss, we could predict with a better than 50% success rate. This possibility should not be allowed to undermine our sensible tendency to assign a probability of a half to the occurrence of heads on the specified toss when we *lack* that microscopic data.

4. <u>A simple system</u>

Here is a very simple example of how objective (frequency or chance) probability can be applied to "unique events". It is essentially due to Reichenbach [1949]. It is a step backward from the discussion of the previous section, but we will regain our insights in the following section.

Let $R = \{r_1, \ldots r_n\}$ be a finite set of potential reference classes, let $P = \{P_1, \ldots, P_k\}$ be a finite set of properties (including such properties as that of being a member of a particular reference class), and let $I = \{i_1, \ldots, i_m\}$ be a set of distinct individuals or individual events.

We can define a language on this basis in the usual way. The reference classes may be of high dimension; predicates may represent complicated relations; individuals may be ordered $t$-tuples of elementary individuals.

Add to this language enough mathematics to do statistics, and define an *item of possible statistical knowledge* to be a sentence of the syntactical form: "$\%(r_j, P_s) = x$", which we read: the proportion of objects in the reference class $r_j$ that have the property $P_s$ is $x$. This is <u>not</u> intended to be a transcription of past frequencies; it is intended to reflect an inference that will often, but not always, be based on the observation of past frequencies.

In general, for the simple case, the following axioms suffice to yield appropriate objective probabilities reflecting what is known -- the evidence -- in $K$.

**A1** If $S$ and $T$ are known to have the same truth-value, then they have the same probability.

This axiom does not require that $S$ and $T$ be equivalent in any strong sense; all that is required is that we know that they have the same truth value.

**A2.1** $R$ is closed under intersection.

**A2.2** If $i \neq j$, $r_i$, $r_j \in R$, then $\sim "r_i = r_j" \in K$.

**A2.3** If $r_i$ $r_j$ then "$r_i$ $r_j$" $\in K$.

Similarly for properties:

**A3.1** $P$ is closed under conjunction and negation.

**A3.2** If $i \neq j$, $P_i$, $P_j \in P$, then $\sim "(x)(P_i(x) \longleftrightarrow P_j(x))" \in K$.

**A3.3** If $(x)(P_i(x) \longrightarrow P_j(x))$, then "$(x)(P_i(x) \longrightarrow P_j(x))$" $\in K$.

A bit of logical closure:

**A4** If "$i_x \in r_y$" $\in K$ and "$i_x \in r_w$" $\in K$, then "$i_x \in r_y$ $r_w$" $\in K$.

To insure that "having the same truth value" generates equivalence classes, we stipulate for any $S$,

**A5.1** If "$S_1 \longleftrightarrow S_2$" $\in K$, then "$S_2 \longleftrightarrow S_1$" $\in K$.

**A5.2** If "$S_1 \longleftrightarrow S_2$" and "$S_2 \longleftrightarrow S_3$" are in $K$, then "$S_1 \longleftrightarrow S_3$" is in $K$.

**A5.3** "$S_1 \longleftrightarrow S_1$" $\in K$.

And finally,

**A6** For every non-mathematical sentence $S$ in our language, there exists a $P_y$ and exactly one $i_x$ such that "$S \longleftrightarrow P_y(i_x)$" $\in K$.

**A7** There exists a model of the sentences in $K$, with "$\%(X, Y)$" construed as the proportion of $X$'s that are $Y$'s.

We can now define the probability of a sentence $S$ relative to a body of knowledge $K$ to be $x$ just in case $S$ is known in $K$ to be equivalent to a sentence of the form "$P_z(i_y)$" -- this is just to say that the biconditional "$S \longleftrightarrow P_z(i_y)$" is in $K$ -- and for some reference class $r_w$ to which $i_y$ is known to belong, "$\%(r_w, P_z) = x$" is in $K$, and, finally, if $r_w'$ is another reference class to which $i_y$ is known to belong, and "$\%(r_w', P_z) \neq x$" is in $K$, then it is known that $r_w$ is included in $r_w'$. (This is just to say that "$r_w \subset r_w'$" $\in K$.) Formally,

D1   $Prob(S, K) = x$ if and only if there are $P_z$, $i_y$, and $r_w$ such that
   (1)  "$S \longleftrightarrow P_z(i_y)$" $\in K$.
   (2)  "$i_y \in r_w$" $\in K$.
   (3)  "$\%(r_w, P_z) = x$" $\in K$.
   (4)  If "$i_y \in r_w'$" $\in K$, and "$\%(r_w', P_z) \neq x$" is in $K$, then "$r_w \subset r_w'$" $\in K$.

Thus $r_w$ is the *smallest* reference class about which we have statistical information to which $i_y$ is known to belong. This is essentially Reichenbach's idea except for the addition of axiom **A1**.

We can generate the probability more clearly by putting the fourth condition as a constraint on a table. Let the first column of the table contain a list of all the reference classes $r_w$ to which $i_y$ is known to belong. Let the second column contain the value of $x(r_w)$ from the corresponding item of possible statistical knowledge: "$\%(r_w, P_z) = x(r_w)$". Work down the table, deleting every row that fails condition (4). (Rule: if $x(r_w) \neq x(r_w')$, delete both rows unless "$r_w \subset r_w'$" is in $K$.) There may be several rows left, but they will all mention the same value of $x$. There *may* be no rows left.

5. <u>Limitations.</u>

This approach deals perfectly reasonably with tosses

153

of coins and the like. But it fails to provide for the case in which we get a probability from <u>approximate</u> knowledge of frequencies. It gives us no probability at all when we know of $i_y$ that it belongs to *two* reference classes, our knowledge about those reference classes doesn't agree, and we don't know that either reference class is included in the other.

The remedy is simple and obvious, but it entails considerable complication. We allow items of possible statistical knowledge to embody *approximate* knowledge. To stick to one syntactical form, let us write

$$\text{"}\%(r_w, P_z) \in [x_1, x_2]\text{"}$$

to mean that the proportion of objects (or the chance of an object) in the reference class $r_w$ having the property $P_z$ is in the closed interval $[x_1, x_2]$.

Suddenly we have statistical knowledge about every property and every reference class: at the very least we will know that the proportion lies in $[0,1]$. And now what do we mean by " $\neq$ "? These changes work: Say that two intervals "differ" if neither is included in the other, replace "$x$" by "$[x_1, x_2]$" throughout **D1,** and rewrite clause (4) of **D1** to say that if $r_w$ and $r_w'$ differ, then $r_w$ is known to be included in $r_w'$:

(4') If "$i_y \in r_w'$" is in $K$ and
"$\%(r_w', P_z) \in [x_1, x_2]$" is in $K$
and $[x_1', x_2']$ differs from $[x_1, x_2]$,
then "$r_w \subset r_w'$" is in $K$.

This definition of probability is still limited -- it turns out that we would like two other relations, in addition to the subset relation, to excuse "difference". And we would like to be able to consider equivalence among statements concerning different individuals. (A general definition along these lines is provided in [1983].) But it is already quite powerful, and has some interesting properties:

(1) *All* probabilities are objective, in the sense that every probability is based on empirical knowledge about frequencies or chances in the world.

(2) *Every* statement in the language *has* a

probability: there is no distinction between statements concerning "repeatable" events and statements concerning "unique" events.

(3) *No a priori* probabilities are required; all probabilities can be based on experience. (But *how* they can be so based is another story.)

(4) Any application of Bayes' theorem that is justified by background knowledge of frequencies will be preserved.

6. Conclusion.

Probabilistic knowledge may be regarded as all of a piece. There is no need to distinguish between "statistical" probabilities that have objective warrant in the world, and "subjective" probabilities that merely reflect our subjective feelings. When we apply our knowledge of statistical facts to individual cases it is the probability of a unique event that is at issue but it is based on some (possibly approximate) statistical knowledge.

*This work has been supported in part by the Signals Warfare Center of the U. S. Army.